\documentclass[letterpaper]{article}
\usepackage{aaai19}
\usepackage{times}
\usepackage{helvet}
\usepackage{courier}
\usepackage[hyphens]{url}
\usepackage{graphicx}
\usepackage{amsmath}
\usepackage{amsfonts}
\usepackage{natbib}
\usepackage{numprint}
\usepackage{graphicx}
\def\reals{\mathbb{R}}
\def\normal{\mathcal{N}}

\title{Newtonian Monte Carlo: single-site MCMC meets second-order gradient methods}
\author{
Nimar S. Arora, Nazanin Khosravani Tehrani, Kinjal Divesh Shah,
Michael Tingley, Yucen Lily Li, \\
{\bf \Large Narjes Torabi, David Noursi,
Sepehr Akhavan Masouleh, Eric Lippert, Erik Meijer} \\
\{nimarora, nazaninkt, kshah97, tingley, yucenli,
ntorabi, dcalifornia, sepehrakhavan, ericlippert, erikm
\}@fb.com\\
   Facebook Inc, 1 Hacker Way, Menlo Park CA 94025
 }
\begin{document}
\maketitle
\begin{abstract}
Single-site Markov Chain Monte Carlo (MCMC) is a variant of MCMC in which a single coordinate in the state space is modified in each step.
Structured relational models are a good candidate for this style of inference.
In the single-site context, second order methods become feasible because the typical cubic costs associated with these methods is now restricted to the dimension of each coordinate.
Our work, which we call Newtonian Monte Carlo (NMC), is a method to improve MCMC convergence by analyzing the first and second order gradients of the target density to determine a suitable proposal density at each point.
Existing first order gradient-based methods suffer from the problem of determining an appropriate step size.
Too small a step size and it will take a large number of steps to converge, while a very large step size will cause it to overshoot the high density region.
NMC is similar to the Newton-Raphson update in optimization where the second order gradient is used to automatically scale the step size in each dimension.
However, our objective is to find a parameterized proposal density rather than the maxima.

As a further improvement on existing first and second order methods, we show that random variables with constrained supports don't need to be transformed before taking a gradient step.
We demonstrate the efficiency of NMC on a number of different domains.
For statistical models where the prior is conjugate to the likelihood, our method recovers the posterior quite trivially in one step.
However, we also show results on fairly large non-conjugate models, where NMC performs better than adaptive first order methods such as NUTS or other inexact scalable inference methods such as Stochastic Variational Inference or bootstrapping.
\end{abstract}

\section{Introduction}
\label{sec:intro}
Markov Chain Monte Carlo (MCMC) methods are often used to generate samples from an unnormalized probability density $\pi(\theta)$ that is easy to evaluate but hard to directly sample.
Such densities arise quite often in Bayesian inference as the posterior of a generative model $p(\theta, Y)$ conditioned on some observations $Y=y$, where $\pi(\theta) = p(\theta,y)$.
The typical setup is to select a {\em proposal} distribution $q(.|\theta)$ that proposes a move of the Markov chain to a new state $\theta^* \sim q(.|\theta)$.
The Metropolis-Hastings acceptance rule is then used to accept or reject this move with probability:
\[\min\left[
    1,
    \frac{\pi(\theta^*) q(\theta|\theta^*)}{\pi(\theta) q(\theta^*|\theta)}
  \right]
.\]

When $\theta \in \reals^k$, a common proposal density is the Gaussian distribution $\normal(\theta, \epsilon^2 I_k)$ centered at $\theta$ with covariance $\epsilon^2 I_k$, where $\epsilon$ is the step size and $I_k$ is the identity matrix defined over $\reals^{k,k}$.
This proposal forms the basis of the so-called Random Walk MCMC (RWM) first proposed in \cite{metropolis1953equation}.

In cases where the target density $\pi(\theta)$ is differentiable, an improvement over the basic RWM method is to propose a new value in the direction of the gradient, as follows:
\[ q(.|\theta) = \normal\left(\theta + \frac{\epsilon^2}{2}\nabla \log \pi(\theta), \epsilon^2 I_k \right).\]
This method is known as Metropolis Adjusted Langevin Algorithm (MALA), and arises from an Euler approximation of a Langevin diffusion process \citep{robert1996exponential}.
MALA has been shown to reduce the number of steps required for convergence to $O(n^{1/3})$ from $O(n)$ for RWM \citep{roberts1998optimal}.
An alternate approach, which also uses the gradient, is to do an $L$-step Euler approximation of Hamiltonian dynamics known as Hamiltonian Monte Carlo \citep{neal1993bayesian}, although it was originally published under the name Hybrid Monte Carlo \citep{duane1987hybrid}.

In HMC the number of steps, $L$, can be learned dynamically by the No-U-Turn Sampler (NUTS) algorithm \citep{hoffman2014no}.
However, in all three of the above algorithms -- RWM, MALA, and HMC -- there is an open problem of selecting the optimal step size.
Normally, the step size is adaptively learned by targeting a desired acceptance rate.
This has the unfortunate effect of picking the same step size for all the dimensions of $\theta$, which forces the step size to accomodate the dimension with the smallest variance as pointed out in \cite{girolami2011riemann}.
The same paper introduces alternate approaches, using Riemann manifold versions of MALA (MMALA) and HMC (RMHMC).
They propose a Riemann manifold using the expected Fisher information matrix plus the negative Hessian of the log-prior as a metric tensor,
$ -E_{y|\theta}\left[ \frac{\partial ^ 2}{\partial \theta^2} log \{ p(y, \theta) \} \right]$,
and proceed to derive the Langevin diffusion equation and Hamiltonian dynamics in this manifold.
The use of the above metric tensor does address the issue of differential scaling in each dimension.
However, the method as presented requires analytic knowledge of the Fisher information matrix. This makes it difficult to design inference techniques in a generic way, and requires derivation on a per-model basis.
A more practical approach involves using the negative Hessian of the log-probability as the metric tensor, $-\frac{\partial ^ 2}{\partial \theta^2} log \{ p(y, \theta) \}$.
However, this encounters the problem that this is not necessarily positive definite throughout the state space.
An alternate approach for scaling the moves in each dimension is to use a preconditioning matrix $M$ \citep{roberts2002langevin} in MALA,
\[q(.|\theta) = \normal\left(\theta + \epsilon^2 M \nabla \log \pi(\theta), \epsilon^2 M \right),\]
also known as the mass matrix in HMC and NUTS, but it's unclear how to compute this.

Another approach is to approximately compute the Hessian \citep{zhang2011quasi} using ideas from quasi-Newton optimization methods such as L-BFGS \citep{nocedal2006numerical}.
This approach and its stochastic variant \citep{simsekli2016stochastic} use a fixed window of previous samples of size $M$ to approximate the Hessian.
However, this makes the chain an order $M$ Markov chain, which introduces considerable complexity in designing the transition kernel in addition to introducing a new parameter $M$.
The key observation in our work is that for single-site methods we only need to compute the Hessian of one coordinate at a time, and this is usually tractable.
The other key observation is that we don't need to always make a Gaussian proposer using the Hessian.
In cases when the coordinate under consideration is a constrained random variable then we can propose from any parameterized density in the same constrained space by matching its curvature.
This approach of curvature-matching to an approximating density allows us to deal with constrained random variables without introducing a transformation such as in Stan \citep{carpenter2017stan}.

In the rest of the paper, we will describe our approach to exploit the curvature of the target density, and show some results on multiple data sets.

\section{Newtonian Monte Carlo}
\label{sec:newtonian}
\subsection{Overview}
\label{sec:newtonian-overview}
This paper introduces the Newtonian Monte Carlo (NMC) technique for sampling from a target distribution via a proposal distribution that incorporates curvature around the current sample location. We wish to choose a proposal distribution that uses second order gradient information in order to closely match the target density. Whereas related MCMC techniques discussed in Section \ref{sec:intro} may utilize second order gradient information, those techniques typically use it only to adjust the step size when simulating steps along the general direction of the target density's gradient.

Our proposed method involves matching the target density to a parameteric density that best explains the current state. We have a library of 2-parameter target densities $F_i$, and simple inference rules such that, given the first and second order gradients, we can solve the
following two equations:
\begin{align*}
\nabla \log \pi(\theta) &= \frac{\partial}{\partial \theta} F_i(\theta ; \alpha_i, \beta_i) \\
\nabla^2 \log \pi(\theta) &= \frac{\partial^2}{\partial \theta^2} F_i(\theta ; \alpha_i, \beta_i),
\end{align*}
to determine $\alpha_i$ and $\beta_i$. For example, in the case of $\theta \in \reals^k$, we use either the multivariate Gaussian or the multivariate Cauchy.
For the former, the update equation leads to the natural proposal,
\[ \normal(\theta - \nabla^2 \log \pi(\theta)^{-1} \nabla \log \pi(\theta), -\nabla^2 \log \pi(\theta)^{-1}). \]
The update term in the mean of this multivariate Gaussian is precisely the update term of the Newton-Raphson Method \citep{newtonsmethod}, which is where NMC gets its name from. However, if the negative Hessian inverse matrix is not positive definite, then the multivariate normal is not defined. In this case, we can instead use the Cauchy proposer as shown by \cite{minka2000beyond}. The full list of estimation methods are enumerated in Section \ref{sec:estimation}.
For example, for positive real values we use a Gamma proposer with parameters,
\begin{align*}
\alpha &= 1 - x^2 \nabla^2 \log \pi(x)\\
\beta &= - x \nabla^2 \log \pi(x) - \nabla \log \pi(x),
\end{align*}
and we don't need a log-transform to an unconstrained space. We rely on generic Tensor libraries such as PyTorch \citep{paszke2017automatic} that make it easy to write statistical models and also automatically compute the gradients. This makes our approach easy to apply to models generically.

In the case of conjugate models, our estimation methods automatically recover the appropriate conditional posterior distribution, such as the ones used in BUGS \citep{spiegelhalter1996bugs}.
However, even in cases of non-conjugacy, our proposal distributions pick out reasonable approximations to the conditional posterior of each variable.

\subsection{Single Site Inference}
\label{sec:newtonian-singlesite}
An important observation related to our method is that we don't need to compute the joint Hessian of all the parameters in the latent space.
Most statistical models with relational structure can be decomposed into multiple latent variables.
This decomposition allows for single site MCMC methods that change the value of one variable at a time.
In this case, we only need to compute the gradient and Hessian of the target density w.r.t. the variable being modified.
Consider a model with $N$ variables each drawn from $R^K$.
The full Hessian is of size $(NK)^2$ and has a cost of $(NK)^3$ to invert.
On the other hand, a single site approach computes $N$ Hessians each of size $K^2$ with a total cost of $NK^3$ to invert.

\section{Estimation Rules}
\label{sec:estimation}
The estimation rules presented here are based on the work of \cite{minka2000beyond}.

\subsection{Unconstrained spaces}
We first consider distributions with the support $\reals^k$.

\subsubsection{Normal Distribution}
The multivariate Normal distribution has the log-density:
\[\text{Normal}(x; \mu, \Sigma) = \text{const}(\mu, \Sigma) - \frac{1}{2} (x - \mu)^T \Sigma^{-1} (x-\mu)\]
Thus,
\begin{align*}
\frac{\partial}{\partial x} \text{Normal}(x; \mu, \Sigma) &=  - \Sigma^{-1} (x - \mu), \text{and}\\
\frac{\partial^2}{\partial x^2} \text{Normal}(x; \mu, \Sigma) &= - \Sigma^{-1}.\\
\end{align*}
This leads to the natural estimation rule:
\begin{align*}
\mu &= x -\nabla^2 \log \pi(x) ^ {-1} \nabla \log \pi(x),\\
\Sigma &= -\nabla^2 \log \pi(x) ^ {-1}. \\
\end{align*}
In case the estimated $\Sigma$ has a negative eigenvalue we set those negative eigenvalues to a very small positive number, and reconstruct $\Sigma$.

\subsubsection{Cauchy Distribution}
The multivariate Cauchy distribution has the log-density:
\[\text{Cauchy}(x; b, A) = \text{const}(b, A) - \log (1 + (x - b)^T A (x - b))\]
Thus,
\begin{align*}
\frac{\partial}{\partial x} \text{Cauchy}(x; b, A) &= \frac{-2 A (x - b)}{1 + (x-b)^T A (x-b)}, \text{and}\\
\frac{\partial^2}{\partial x^2}  \text{Cauchy}(x; b, A) &=
  \frac{-2 A }{1 + (x-b)^T A (x-b)} \\
  &+ \frac{4 A (x - b)(x-b)^T A}{(1 + (x-b)^T A (x-b))^2}.
\end{align*}
Noting that the second term above is the outer product of the first gradient leads to the following estimation rules:
\begin{align*}
b &=  x \ -\\
& \left( \nabla^2 \log \pi(x) -  \nabla \log \pi(x)  \nabla \log \pi(x)^T \right)^{-1} \nabla \log \pi(x),\\
s &= \nabla \log \pi(x)^T \left( \nabla^2 \log \pi(x) \right)^{-1} \nabla \log \pi(x), \\
A &= \left( \nabla^2 \log \pi(x) -  \nabla \log \pi(x)  \nabla \log \pi(x)^T \right) \frac{s-1}{2-s}.
\end{align*}

\subsection{Constrained Spaces}

\subsubsection{Half-Space}
We use the Gamma distribution for proposing values for variables which lie on any half-space constrained distribution, i.e. $\reals^+$. The Gamma distribution has the log-density:
\begin{align*}
\text{Gamma}(x; \alpha, \beta) &= \text{const}(\alpha, \beta) + (\alpha-1) \log x - \beta x.
\intertext{Thus,}
\frac{\partial}{\partial x}  \text{Gamma}(x; \alpha, \beta) &=  \frac{\alpha-1}{x} -  \beta,\\
\frac{\partial^2}{\partial x^2}  \text{Gamma}(x; \alpha, \beta) &=  -\frac{\alpha-1}{x^2}.
\end{align*}
Which leads to the estimation rules:
\begin{align*}
\alpha &= 1 - x^2 \nabla^2 \log \pi(x),\\
\beta &= - x \nabla^2 \log \pi(x) - \nabla \log \pi(x), \\
\end{align*}

\subsubsection{Simplexes}
The K-simplexes refers to the set $\{ x \in {\reals^+}^K | \sum_{i=1}^K x_i  = 1\}$.
We use the Dirichlet distribution to propose random variables with this support.
The log-density of the Dirichlet is given by,
\begin{align*}
\text{Dir}(x; \alpha) &= \text{const}(\alpha) +  \sum_{i=1}^{K} (\alpha_i-1) \log (x_i).
\intertext{We consider the modified density, which includes the simplex constraint,}
\text{Dir}(x; \alpha) &= \text{const}(\alpha) +  \sum_{i=1}^{K} (\alpha_i-1) \log \frac{x_i}{\sum_{j=1}^K x_j}.
\intertext{Thus,}
\frac{\partial}{\partial x_i} \text{Dir}(x; \alpha) &= \frac{(\alpha_i-1)}{x_i} -  \frac{\sum_{j=1}^K (\alpha_j-1)}{\sum_{j=1}^K x_j} , \text{and}\\
\frac{\partial^2}{\partial x_i \partial x_l} \text{Dir}(x; \alpha) &= -\delta_{il} \frac{(\alpha_i-1)}{x_i^2} + \frac{\sum_{j=1}^K (\alpha_j-1)}{(\sum_{j=1}^K x_j)^2}.
\end{align*}
Which leads to  the following estimation rule,
\begin{align*}
\alpha_i &= 1 - x_i^2 \left(\nabla_{ii}^2 \log \pi(x) - \max_{j\ne i} \nabla_{ij}^2 \log \pi(x)\right),
\end{align*}
for $i=1\ldots K$.

\section{Experiments}
\label{sec:experiments}

\subsection{Models}
\label{sec:models}

\subsubsection{Neal's Funnel}
\label{sec:funnel}
\begin{figure}[htb]
\centering
\includegraphics[width=0.65\columnwidth]{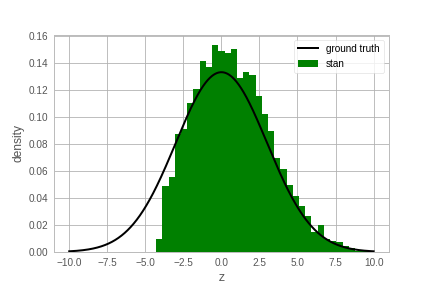}
\caption{Posterior marginal of $z$ in Neal's funnel after $\numprint{10000}$ Stan samples}
\label{fig:stan-funnel}
\end{figure}

\begin{figure}[htb]
\centering
\includegraphics[width=0.65\columnwidth]{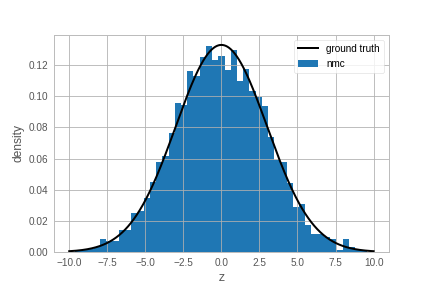}
\caption{Posterior marginal of $z$ in Neal's funnel after $\numprint{10000}$ NMC samples}
\label{fig:nmc-funnel}
\end{figure}

We first consider a toy model, which is considered difficult for MCMC methods.
The model was first proposed in \cite{neal2003slice} and has been since called Neal's Funnel.
The following equations define the joint density of the model, which is deceptively simple.
\begin{align*}
z &\sim \normal(0, 3) \\
x &\sim \normal(0, e^{\frac{z}{2}})
\end{align*}
The difficulty for inference arises when we try to sample values of $(x, z)$ for increasingly negative $z$. Since the scale of $x$ varies exponentially with $z$ it is hard to learn a good scale.
Indeed Figure~\ref{fig:stan-funnel} shows that Stan, which uses NUTS, has a hard time sampling from this distribution as highlighted by the posterior marginal of $z$.
On the other hand NMC, which effectively computes the scale dynamically at each point has no difficulty in generating good samples, as shown in Figure~\ref{fig:nmc-funnel}.

\subsubsection{Bayesian Logistic Regression}
Next we consider the Bayesian Logistic Regression model that is commonly used in machine learning. The model is defined as follows:
\label{sec:blr}
\begin{align*}
\alpha &\sim \normal(0, 10, \text{size}=1),\\
\beta &\sim \normal(0, 2.5, \text{size}=K),\\
X_i &\sim \normal(0, 10, \text{size}=K) \quad \forall i \in 1..N \\
\mu_i &= \alpha + {X_i}^T \beta \quad \forall i \in 1..N\\
Y_i &\sim \text{Bernoulli}(\text{logit}=\mu_i) \quad \forall i \in 1..N.
\end{align*}
From this model we generate samples of $\alpha$, $\beta$, $X_i$ and $Y_i$ for some given $N$ and $K$.
Half of the $X,Y$ samples are given to the inference method to produce posterior samples of $\alpha$ and $\beta$.
The held out values of $X, Y$ are used to evaluate the predictive log-likelihood, which is averaged over the posterior samples.
In this and the rest of the models, we draw exactly $\numprint{1000}$ samples using each method.
For single-site methods a sample includes an update to each coordinate.

\begin{figure*}[htb]
\centering
\includegraphics[width=0.65\textwidth]{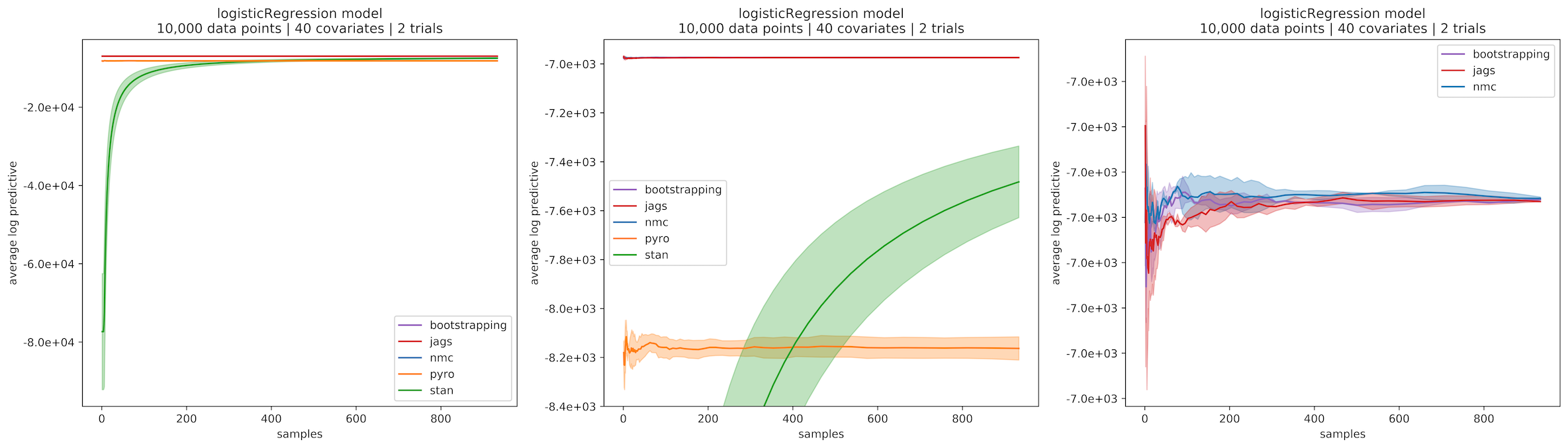}
\caption{Results for logistic regression with $N=\numprint{20000}$ and $K=\numprint{40}$.
The leftmost figure shows all the methods together, and the subsequent figures show a zoomed in view.}
\label{fig:blr-20k}
\end{figure*}

\begin{figure*}[htb]
\centering
\includegraphics[width=0.65\textwidth]{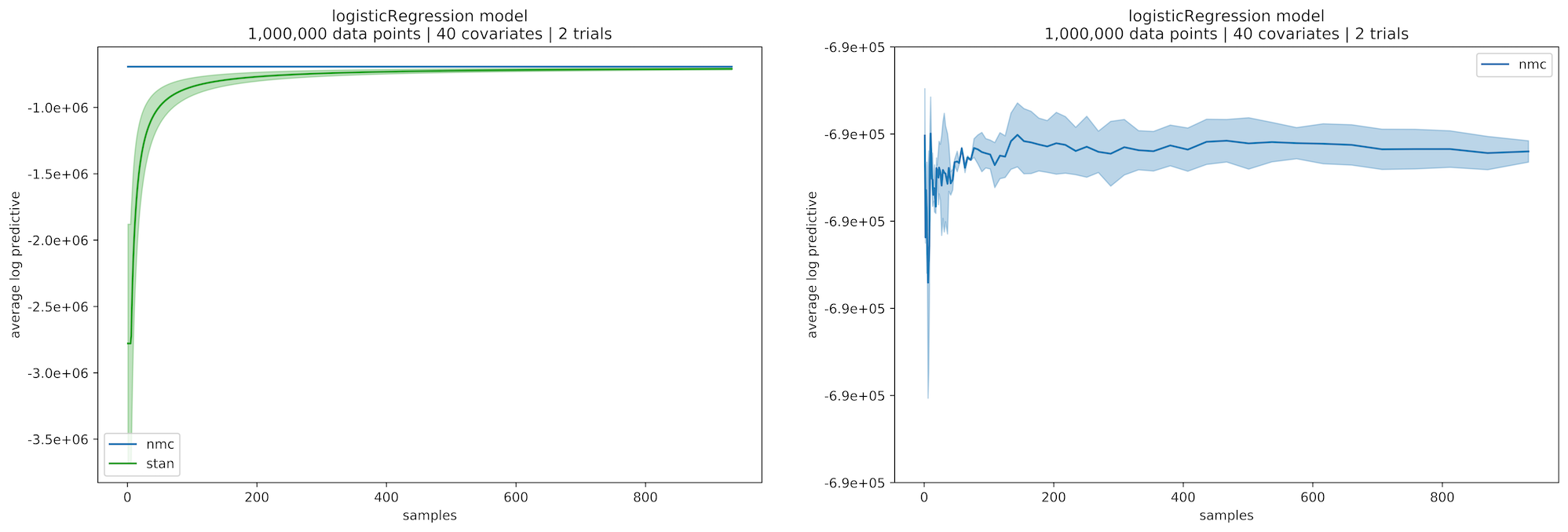}
\caption{Results for logistic regression with $N=\numprint{2000000}$ and $K=\numprint{40}$.
The figure on the right shows a zoomed in view.}
\label{fig:blr-20M}
\end{figure*}

Figure~\ref{fig:blr-20k} shows results for $N=\numprint{20000}$ using a variety of methods. Table~\ref{table:blr-runtimes} shows the run times to produce $\numprint{1000}$ samples, plus the number of samples required to achieve convergence.
We have defined samples to convergence to be the number of samples it takes for the predictive log-likelihood to stabilize to within $1\%$ of the final value for the method.
Figure~\ref{fig:blr-20M} shows results for $N=\numprint{2000000}$, where we only compare two of the methods since the other methods were too slow for such a large data set.

This model has a nice log concave posterior which is quite easy for MCMC inference and hence both JAGS and NMC converge in $1$ sample.
Stan, which uses NUTS, does take a bit longer to converge because it has to learn the optimum scale in each of the $K$ dimensions of $\beta$.
Bootstrapping-based approaches are often used for this model, but they do appear to incur additional cost of re-training the model for each bootstrap sample.
Finally, we also used Stochastic Variational Inference as implemented in Pyro \citep{bingham2019pyro}, but it doesn't appear from this example that the loss of accuracy of using variational inference is worthwhile.
In this model, NMC seems to be the fastest both in terms of time per samples and samples to convergence.

\begin{table}[htb]
\caption{Runtimes for Bayesian Logistic Regression.}\smallskip
\centering
\resizebox{.95\columnwidth}{!}{
\smallskip\begin{tabular}{l|l|l|l}
Method & N & Time (seconds) & Samples to convergence\\
\hline
NMC & 20K & \numprint{18} & \numprint{1} \\
Stan & 20K & \numprint{41} & \numprint{616} \\
JAGS & 20K & \numprint{2440} & \numprint{1} \\
Boostrapping & 20K & \numprint{50} & \numprint{1} \\
Pyro & 20K & \numprint{3024} & \numprint{6} \\
\hline
NMC & 20M & 1030 & 1\\
Stan & 20M & 4900 & 380\\
\end{tabular}
}
\label{table:blr-runtimes}
\end{table}

\subsubsection{Robust Regression}
Robust regression is a regression model in which an error distribution with a much wider tail than Gaussian such as the Student's t distribution is used to model data with outliers.
We use the following model:
\label{sec:robust}
\begin{align*}
\nu &\sim \text{Gamma}(2,\ 0.1)\\
\sigma &\sim \text{Exponential}(\sigma_{mean})\\
\alpha &\sim \text{Normal}(0,\ \sigma=\alpha_{scale})\\
\beta &\sim \text{Normal}(\beta_\text{loc},\ \sigma=\beta_{scale},\ \text{size}=K)\\
X_i &\sim \text{Normal}(0,\ 10,\ \text{size}=K) \quad \forall i \in 1\ldots N\\
\mu_i &= \alpha + \beta^T x  \quad \forall i \in 1\ldots N\\
Y_i &\sim \text{Student-T}(\nu,\ \mu_i,\ \sigma) \quad \forall i \in 1\ldots N\\
\end{align*}
As before we generate samples from the model of all the variables including $N$ values of $X_i$ and $Y_i$.
Half of the generated samples are given to the inference algorithm and the other half are used for evaluating the posterior.

\begin{figure*}[htb]
\centering
\includegraphics[width=0.65\textwidth]{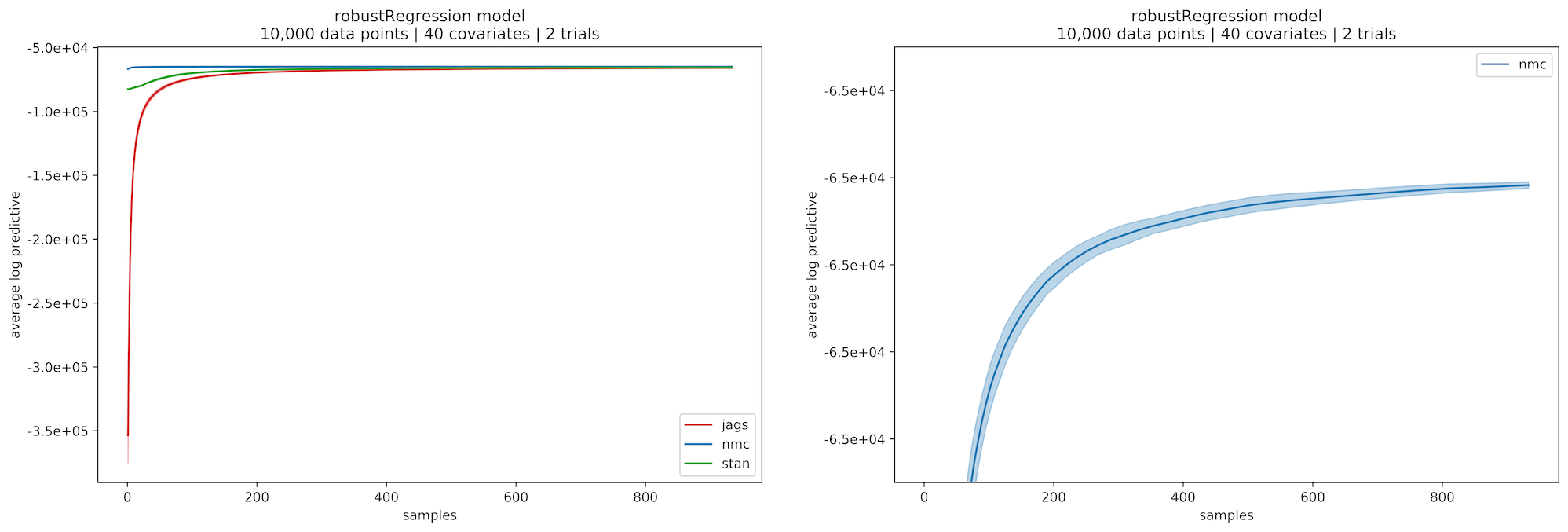}
\caption{Results for robust regression with $N=\numprint{20000}$ and $K=\numprint{40}$.
The figure on the right shows a zoomed in view.}
\label{fig:robust-20k}
\end{figure*}

\begin{figure*}[htb]
\centering
\includegraphics[width=0.65\textwidth]{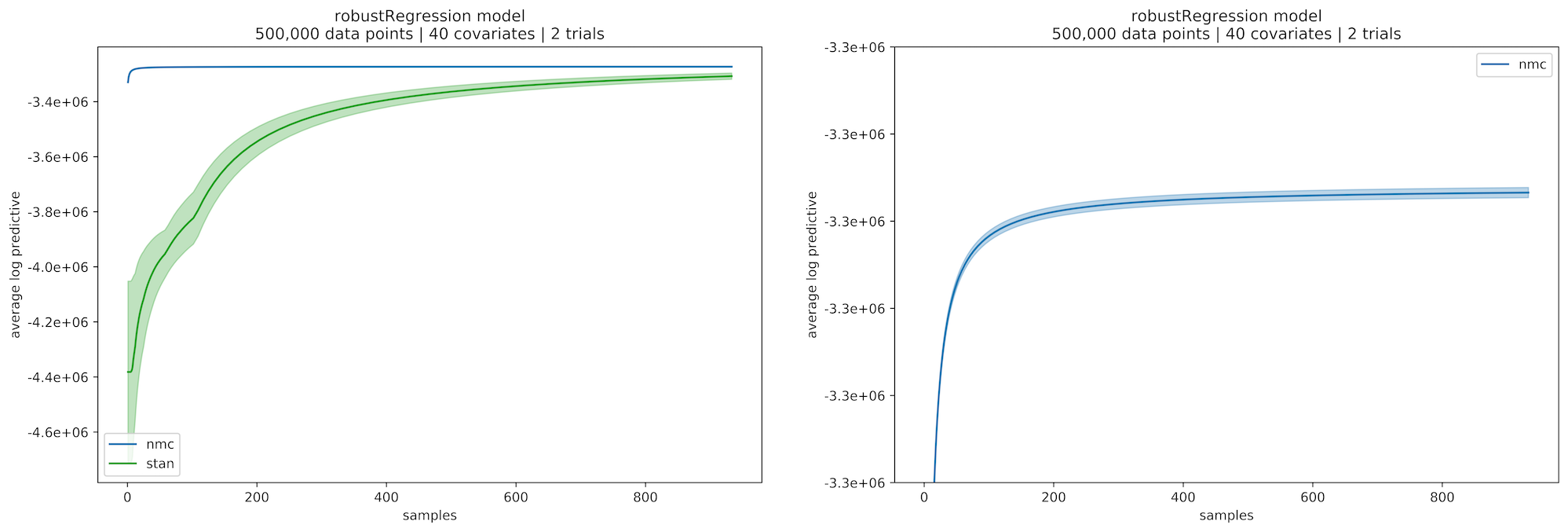}
\caption{Results for robust regression with $N=\numprint{1000000}$ and $K=\numprint{40}$.
The figure on the right shows a zoomed in view.}
\label{fig:robust-1M}
\end{figure*}

This model is not log-concave because of the Student's t distribution, and as a result JAGS takes much longer to converge as shown in Figure~\ref{fig:robust-20k} and and Table~\ref{table:robust-runtimes}.
We also ran an experiment for much larger $N$, Figure~\ref{fig:robust-1M}, where we left out JAGS because it was too slow.
On this model, NMC converges much faster than Stan using merely $3$ samples to converge for the larger data set, and with much faster runtimes as well.

\begin{table}[htb]
\caption{Runtimes for Robust Regression.}\smallskip
\centering
\resizebox{.95\columnwidth}{!}{
\smallskip\begin{tabular}{l|l|l|l}
Method & N & Time (seconds) & Samples to convergence\\
\hline
NMC & 20K & \numprint{68} & \numprint{8} \\
Stan & 20K & \numprint{39} & \numprint{407} \\
JAGS & 20K & \numprint{967} & \numprint{537} \\
\hline
NMC & 1M & \numprint{1777} & \numprint{3} \\
Stan & 1M & \numprint{3500} & \numprint{812} \\
\end{tabular}
}
\label{table:robust-runtimes}
\end{table}

\subsubsection{Annotation Model}
\label{sec:annotation}
Our final model has a lot more relational structure than the regression models above.
This is a slightly modified version of the model presented in \cite{passonneau2014benefits} and \cite{dawid1979maximum}, and is designed to compute the true labels of items given noisy crowd-sourced labels.
There are $N$ items, $K$ labelers, and each item could be one of $C$ categories.
Each item $i$ is labeled by a set $J_i$ of labelers. Such that the size of $J_i$ is sampled randomly, and each labeler in $J_i$ is drawn uniformly without replacement from the set of all labelers.
$z_i$ is the true label for item $i$ and $y_{ij}$ is the label provided to item $i$ by labeler $j$.
Each labeler $l$ has a confusion matrix $\theta_l$ such that $\theta_{lmn}$ is the probability that an item with true class $m$ is labeled $n$ by $l$.
\begin{align*}
\pi &\sim \text{Dirichlet}\left(\frac{1}{C}, \ldots, \frac{1}{C}\right)\\
z_i &\sim \text{Categorical}(\pi)\quad \forall i \in 1\ldots N\\
\theta_{lm} &\sim \text{Dirichlet}(\alpha_m)\quad \forall l \in 1\ldots K,\ m \in 1 \ldots C\\
|J_i| &\sim \text{Poisson}(J_\text{loc})\\
l \in J_i &\sim \text{Uniform}(1\ldots K) \quad \text{without replacement}\\
y_{il} &\sim \text{Categorical}(\theta_{l z_i})\quad \forall l \in J_i\\
\end{align*}
Here $\alpha_m \in {\reals^+}^C$. We set $\alpha_{mn} = \gamma\cdot \rho$ if $m=n$ and $\alpha_{mn} = \gamma \cdot (1 - \rho)\cdot \frac{1}{C-1}$ if $m \ne n$.
Where $\gamma$ is the concentration and $\rho$ is the {\em a-priori} correctness of the labelers.
In this model, $Y_{il}$ and $J_i$ are observed.

In our experiments, we fixed $K=100$, $C=3$, $J_\text{loc}=2.5$, $\gamma=10$, and $\rho=0.5$.
As before we generated data for different sizes of $N$ and used a random partition of the data for inference and evaluation.
Since Stan doesn't support discrete variables such as $z_i$ above, we had to analytically integrate\footnote{This analytical integration is known as marginalization by Stan users} the $z$'s out of the model.
For the purpose of evaluation, since Stan doesn't give us samples of $z$, we integrate over these variables to compute the predictive likelihood, which gives a disadvantage to methods such as JAGS and NMC where the samples depend on a specific value of $z$.

\begin{figure}[htb]
\centering
\includegraphics[width=0.65\columnwidth]{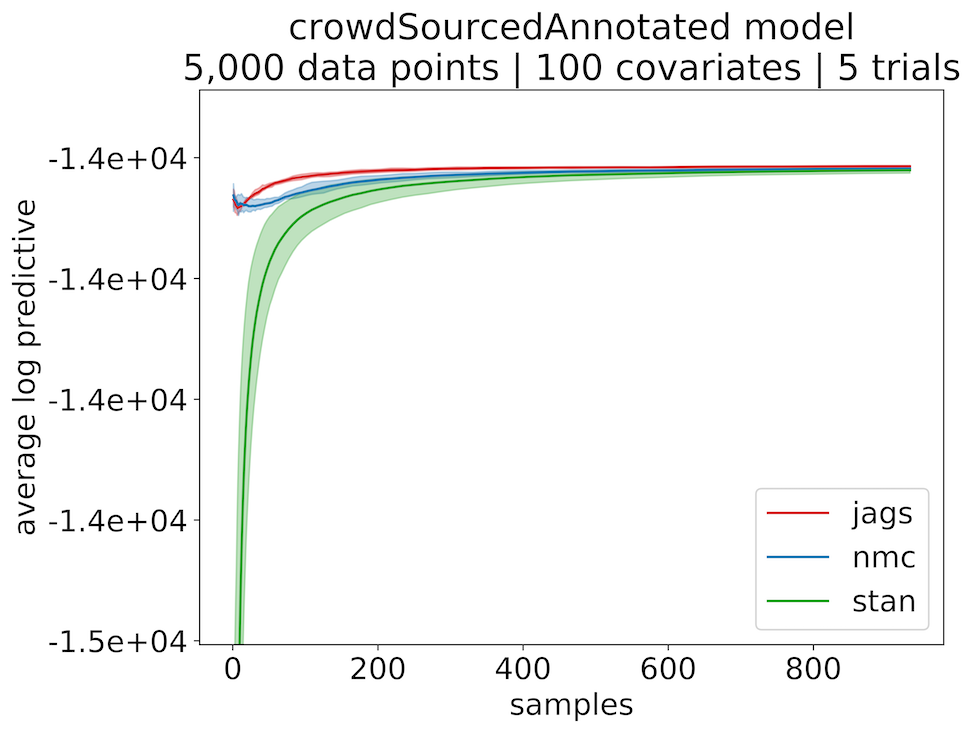}
\caption{Results for annotation model with $N=\numprint{10000}$ and $K$=100.}
\label{fig:annotation-10k}
\end{figure}

\begin{figure}[htb]
\centering
\includegraphics[width=0.65\columnwidth]{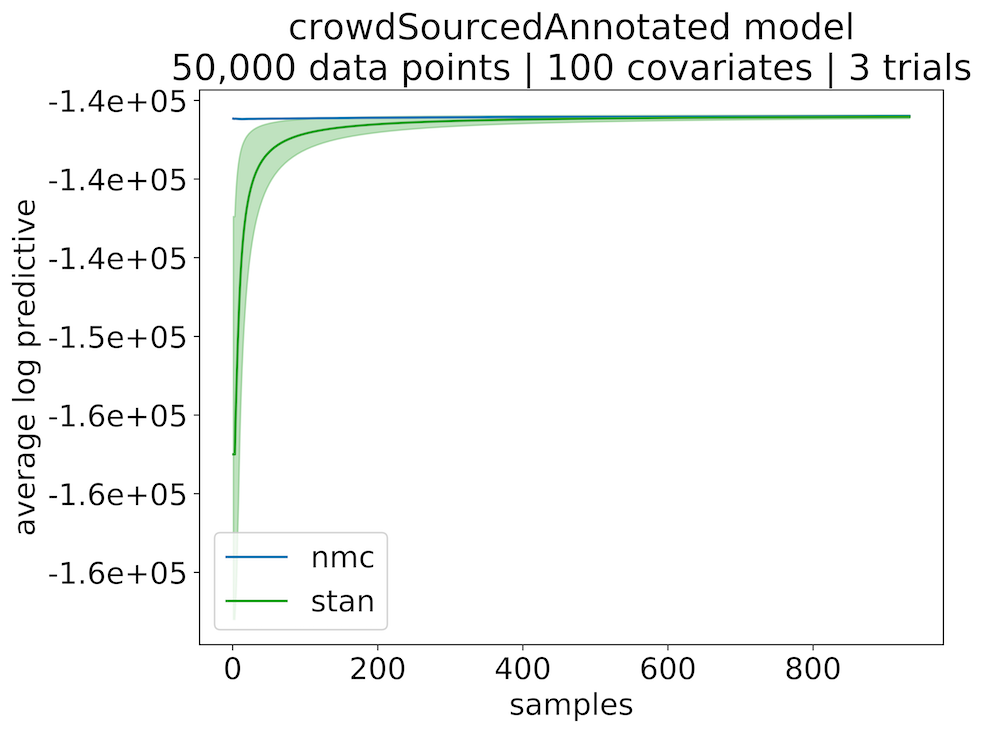}
\caption{Results for annotation model with $N=\numprint{100000}$ and $K$=100.}
\label{fig:annotation-100k}
\end{figure}

\begin{figure}[htb]
\centering
\includegraphics[width=0.65\columnwidth]{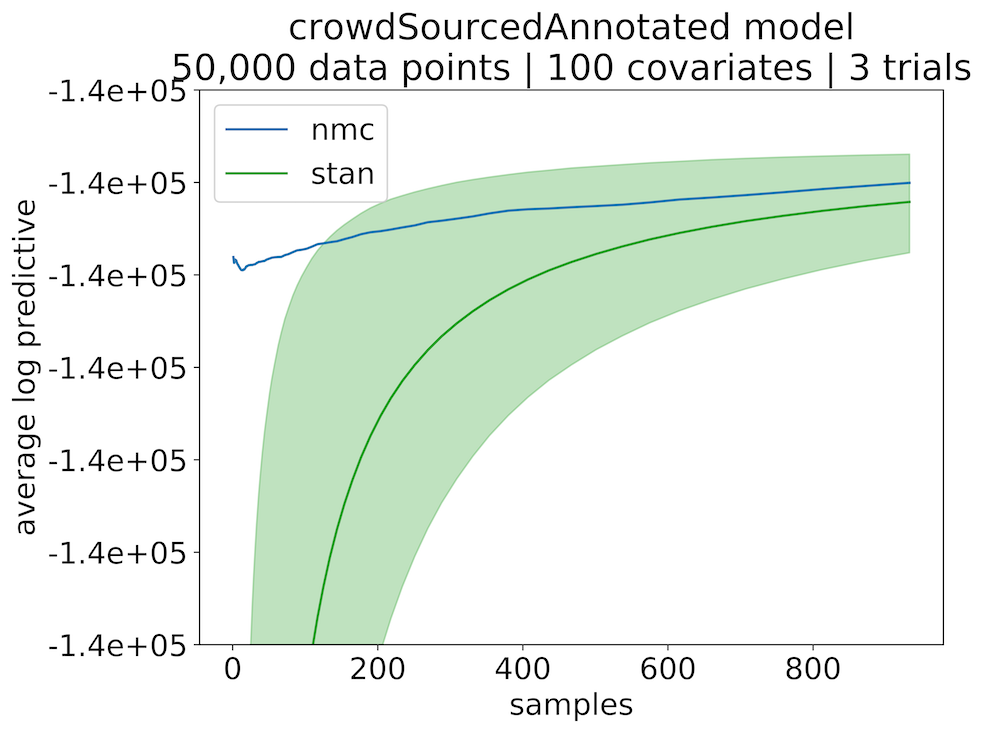}
\caption{Zoomed in view for annotation model with $N=\numprint{100000}$ and $K$=100.}
\label{fig:annotation-100k-2}
\end{figure}

In this model each random variable has a conjugate conditional posterior, and since JAGS is designed to exploit conjugacy it really shines here.
Unfortunately, the version of JAGS that we used kept crashing on larger data sets.
The results for $N=\numprint{10000}$ are shown in Figure~\ref{fig:annotation-10k} and run times are in Table~\ref{table:annotation-runtimes}.
NMC exploits the relational structure in this model to use single-site inference on each of the random variables such as $\theta_{lm}$ individually rather than the entire $\theta$ jointly as in Stan.
As such NMC is easily able to keep up with JAGS in terms of number of samples and is only a factor of \numprint{2.5} slower on the small data set.
On the larger data set, Figure~\ref{fig:annotation-100k}~and~\ref{fig:annotation-100k-2}, NMC is nearly \numprint{7} times faster than Stan.

\begin{table}[htb]
\caption{Runtimes for Annotation Model.}\smallskip
\centering
\resizebox{.95\columnwidth}{!}{
\smallskip\begin{tabular}{l|l|l|l}
Method & N & Time (seconds) & Samples to convergence\\
\hline
NMC & 10K & \numprint{61} & \numprint{1} \\
Stan & 10K & \numprint{387} & \numprint{80} \\
JAGS & 10K & \numprint{31} & \numprint{1} \\
\hline
NMC & 100K & \numprint{410} & \numprint{1} \\
Stan & 100K & \numprint{5294} & \numprint{77} \\
\end{tabular}
}
\label{table:annotation-runtimes}
\end{table}

\section{Conclusion}

We have presented a new MCMC method that uses the first and second gradients of the target density for each coordinate in the latent state space to determine an appropriate proposal distribution.
The method is shown to perform better than the existing state of the art NUTS implementation without requiring an adaptive phase or tuning of inference hyper-parameters.

\clearpage

\bibliographystyle{aaai}
\bibliography{bibfile}

\begin{thebibliography}{}

\bibitem[\protect\citeauthoryear{Bingham \bgroup et al\mbox.\egroup
  }{2019}]{bingham2019pyro}
Bingham, E.; Chen, J.~P.; Jankowiak, M.; Obermeyer, F.; Pradhan, N.;
  Karaletsos, T.; Singh, R.; Szerlip, P.; Horsfall, P.; and Goodman, N.~D.
\newblock 2019.
\newblock Pyro: Deep universal probabilistic programming.
\newblock {\em The Journal of Machine Learning Research} 20(1):973--978.

\bibitem[\protect\citeauthoryear{Carpenter \bgroup et al\mbox.\egroup
  }{2017}]{carpenter2017stan}
Carpenter, B.; Gelman, A.; Hoffman, M.~D.; Lee, D.; Goodrich, B.; Betancourt,
  M.; Brubaker, M.; Guo, J.; Li, P.; and Riddell, A.
\newblock 2017.
\newblock Stan: A probabilistic programming language.
\newblock {\em Journal of statistical software} 76(1).

\bibitem[\protect\citeauthoryear{Dawid and Skene}{1979}]{dawid1979maximum}
Dawid, A.~P., and Skene, A.~M.
\newblock 1979.
\newblock Maximum likelihood estimation of observer error-rates using the em
  algorithm.
\newblock {\em Journal of the Royal Statistical Society: Series C (Applied
  Statistics)} 28(1):20--28.

\bibitem[\protect\citeauthoryear{Duane \bgroup et al\mbox.\egroup
  }{1987}]{duane1987hybrid}
Duane, S.; Kennedy, A.~D.; Pendleton, B.~J.; and Roweth, D.
\newblock 1987.
\newblock Hybrid {M}onte {C}arlo.
\newblock {\em Physics letters B} 195(2):216--222.

\bibitem[\protect\citeauthoryear{Girolami and
  Calderhead}{2011}]{girolami2011riemann}
Girolami, M., and Calderhead, B.
\newblock 2011.
\newblock Riemann manifold {L}angevin and {H}amiltonian {M}onte {C}arlo
  methods.
\newblock {\em Journal of the Royal Statistical Society: Series B (Statistical
  Methodology)} 73(2):123--214.

\bibitem[\protect\citeauthoryear{Hoffman and Gelman}{2014}]{hoffman2014no}
Hoffman, M.~D., and Gelman, A.
\newblock 2014.
\newblock The {N}o-{U}-{T}urn sampler: adaptively setting path lengths in
  {H}amiltonian {M}onte {C}arlo.
\newblock {\em Journal of Machine Learning Research} 15(1):1593--1623.

\bibitem[\protect\citeauthoryear{Metropolis \bgroup et al\mbox.\egroup
  }{1953}]{metropolis1953equation}
Metropolis, N.; Rosenbluth, A.~W.; Rosenbluth, M.~N.; Teller, A.~H.; and
  Teller, E.
\newblock 1953.
\newblock Equation of state calculations by fast computing machines.
\newblock {\em The journal of chemical physics} 21(6):1087--1092.

\bibitem[\protect\citeauthoryear{Minka}{2000}]{minka2000beyond}
Minka, T.~P.
\newblock 2000.
\newblock Beyond {N}ewton’s method.

\bibitem[\protect\citeauthoryear{Neal and others}{2003}]{neal2003slice}
Neal, R.~M., et~al.
\newblock 2003.
\newblock Slice sampling.
\newblock {\em The annals of statistics} 31(3):705--767.

\bibitem[\protect\citeauthoryear{Neal}{1993}]{neal1993bayesian}
Neal, R.~M.
\newblock 1993.
\newblock Bayesian learning via stochastic dynamics.
\newblock In {\em Advances in neural information processing systems},
  475--482.

\bibitem[\protect\citeauthoryear{Nocedal and
  Wright}{2006}]{nocedal2006numerical}
Nocedal, J., and Wright, S.~J.
\newblock 2006.
\newblock Numerical optimization second edition.
\newblock {\em Numerical optimization}  497--528.

\bibitem[\protect\citeauthoryear{Passonneau and
  Carpenter}{2014}]{passonneau2014benefits}
Passonneau, R.~J., and Carpenter, B.
\newblock 2014.
\newblock The benefits of a model of annotation.
\newblock {\em Transactions of the Association for Computational Linguistics}
  2:311--326.

\bibitem[\protect\citeauthoryear{Paszke \bgroup et al\mbox.\egroup
  }{2017}]{paszke2017automatic}
Paszke, A.; Gross, S.; Chintala, S.; Chanan, G.; Yang, E.; DeVito, Z.; Lin, Z.;
  Desmaison, A.; Antiga, L.; and Lerer, A.
\newblock 2017.
\newblock Automatic differentiation in pytorch.

\bibitem[\protect\citeauthoryear{Robert and
  Tweedie}{1996}]{robert1996exponential}
Robert, G., and Tweedie, R.
\newblock 1996.
\newblock Exponential convergence of {L}angevin diffusions and their discrete
  approximation.
\newblock {\em Bernoulli} 2:341--363.

\bibitem[\protect\citeauthoryear{Roberts and
  Rosenthal}{1998}]{roberts1998optimal}
Roberts, G.~O., and Rosenthal, J.~S.
\newblock 1998.
\newblock Optimal scaling of discrete approximations to {L}angevin diffusions.
\newblock {\em Journal of the Royal Statistical Society: Series B (Statistical
  Methodology)} 60(1):255--268.

\bibitem[\protect\citeauthoryear{Roberts and
  Stramer}{2002}]{roberts2002langevin}
Roberts, G.~O., and Stramer, O.
\newblock 2002.
\newblock Langevin diffusions and {M}etropolis-{H}astings algorithms.
\newblock {\em Methodology and computing in applied probability} 4(4):337--357.

\bibitem[\protect\citeauthoryear{Simsekli \bgroup et al\mbox.\egroup
  }{2016}]{simsekli2016stochastic}
Simsekli, U.; Badeau, R.; Cemgil, T.; and Richard, G.
\newblock 2016.
\newblock Stochastic quasi-newton langevin monte carlo.
\newblock In {\em International Conference on Machine Learning (ICML)}.

\bibitem[\protect\citeauthoryear{Spiegelhalter \bgroup et al\mbox.\egroup
  }{1996}]{spiegelhalter1996bugs}
Spiegelhalter, D.; Thomas, A.; Best, N.; and Gilks, W.
\newblock 1996.
\newblock {BUGS} 0.5: Bayesian inference using {G}ibbs sampling manual (version
  ii).
\newblock {\em MRC Biostatistics Unit, Institute of Public Health, Cambridge,
  UK}  1--59.

\bibitem[\protect\citeauthoryear{Whittaker and Robinson}{1967}]{newtonsmethod}
Whittaker, E.~T., and Robinson, G.
\newblock 1967.
\newblock The newton-rhapson method.
\newblock {\em The Calculus of Observations; a Treatise on Numerical
  Mathematics, 4th ed}  84–87.

\bibitem[\protect\citeauthoryear{Zhang and Sutton}{2011}]{zhang2011quasi}
Zhang, Y., and Sutton, C.~A.
\newblock 2011.
\newblock Quasi-newton methods for markov chain monte carlo.
\newblock In {\em Advances in Neural Information Processing Systems},
  2393--2401.

\end{thebibliography}
\end{document}